\documentclass[runningheads]{llncs}
\usepackage[T1]{fontenc}

\usepackage{graphicx}
\usepackage{booktabs}
\usepackage{amsmath}
\usepackage{multirow}
\usepackage{array}

\usepackage{etoolbox}

\newcommand{\repthanks}[1]{\textsuperscript{\ref{#1}}}
\makeatletter
\patchcmd{\maketitle}
  {\def\thanks}
  {\let\repthanks\repthanksunskip\def\thanks}
  {}{}
\patchcmd{\@maketitle}
  {\def\thanks}
  {\let\repthanks\@gobble\def\thanks}
  {}{}
\newcommand\repthanksunskip[1]{\unskip{}}
\makeatother

\begin{document}
\title{PRSM: A Measure to Evaluate CLIP's Robustness Against Paraphrases}
\titlerunning{A Measure to Evaluate CLIP's Robustness Against Paraphrases}

\author{Udo Schlegel\thanks{Equal contribution.\protect\label{X}}\inst{1,2}\orcidID{0000-0002-8266-0162} \and
Franziska Weeber\repthanks{X}\inst{3}\orcidID{0009-0002-6447-8671} \and
Jian Lan\inst{1,2} \and
Thomas Seidl\inst{1,2}\orcidID{0000-0002-4861-1412}
}
\authorrunning{Schlegel et al.}
%
\institute{Ludwig-Maximilians-University Munich (LMU), Germany \and
Munich Center for Machine Learning (MCML), Germany\\
\email{\{schlegel,lan,seidl\}@dbs.ifi.lmu.de}\and
University of Stuttgart, Germany\\
\email{franziska.weeber@ims.uni-stuttgart.de}}
\maketitle              
\begin{abstract}
Contrastive Language-Image Pre-training (CLIP) is a widely used multimodal model that aligns text and image representations through large-scale training. 
While it performs strongly on zero-shot and few-shot tasks, its robustness to linguistic variation, particularly paraphrasing, remains underexplored. 
Paraphrase robustness is essential for reliable deployment, especially in socially sensitive contexts where inconsistent representations can amplify demographic biases. 
In this paper, we introduce the Paraphrase Ranking Stability Metric (PRSM), a novel measure for quantifying CLIP's sensitivity to paraphrased queries. 
Using the Social Counterfactuals dataset, a benchmark designed to reveal social and demographic biases, we empirically assess CLIP's stability under paraphrastic variation, examine the interaction between paraphrase robustness and gender, and discuss implications for fairness and equitable deployment of multimodal systems. 
Our analysis reveals that robustness varies across paraphrasing strategies, with subtle yet consistent differences observed between male- and female-associated queries.

\end{abstract}
\section{Introduction}

Natural language often expresses the same meaning through multiple paraphrases, e.g., “a woman carrying groceries” vs. “a lady holding shopping bags.” 
Prior work shows that both language and vision-language models are sensitive to such variations, even when semantics remain unchanged~\cite{ceron_prompt_2024,wang_adversarial_2021,jung_flex_2025}.
We define \emph{paraphrase robustness} as the ability of a model to produce stable outputs under such rephrasings. 
For vision-language models like CLIP~\cite{radford_learning_2021}, this sensitivity can lead to inconsistent retrieval in downstream applications: Semantically equivalent queries may yield different images. 
Robustness is essential for reliable retrieval and fairness, particularly in socially sensitive settings where instability can amplify demographic biases~\cite{jung_flex_2025}, or for video retrieval tasks (e.g., Video Browser Showdown~\cite{rossetto_interactive_2020}), where minor lexical changes should not alter results.

Existing evaluations of paraphrastic variation largely focus on adversarial perturbations or task-specific accuracy, offering limited insight into retrieval ranking consistency.
To address this gap, we make three contributions. 
First, we propose the \textbf{Paraphrase Ranking Stability Metric (PRSM)}, a general-purpose evaluation measure that captures both global ranking preservation and top-$k$ overlap, enabling fine-grained analysis of retrieval robustness. 
Second, we conduct a case study of CLIP’s paraphrase robustness against three different kinds of paraphrase strategies on the Social Counterfactuals dataset ~\cite{howard_socialcounterfactuals_2024}, covering over 170k query–result pairs with controlled variations in gender and other demographic attributes. 
Third, we analyze how paraphrase sensitivity interacts with gender, showing that while local stability is moderately high, global instability persists and small but systematic differences exist between male- and female-associated queries. 
Together, these findings demonstrate that paraphrase sensitivity is not only a challenge for semantic consistency but also a potential amplifier of bias, underscoring the need for paraphrase-invariant training and bias-aware evaluation in multimodal systems.

\section{Related Work}

Robustness and bias have been studied in language, vision, and multimodal models, each exposing distinct but related vulnerabilities. 
In retrieval tasks, vision-language systems like CLIP are especially sensitive since instability in text embeddings under paraphrasing directly impacts cross-modal alignment. 
We therefore review prior work on (i) vision-language models, (ii) paraphrase robustness in language models, and (iii) robustness in vision models, before motivating our proposed Paraphrase Ranking Stability Metric.

\textbf{Vision-Language Models --}
CLIP~\cite{radford_learning_2021} aligns visual and textual modalities by training on large-scale image-text pairs. Its zero-shot success has spurred interest in applications ranging from retrieval to multimodal reasoning. 
Prior research has identified vulnerabilities in CLIP, including adversarial perturbations~\cite{wortsman_robust_2022}, fairness concerns~\cite{luo_fairclip_2024}, and compositionality limitations~\cite{yuksekgonul_when_2022}. 
The Social Counterfactuals dataset~\cite{howard_socialcounterfactuals_2024} extends this line of inquiry by probing models under counterfactual replacements and paraphrastic variation, exposing the interaction of linguistic generalization and social bias. 
However, CLIP’s specific behavior on paraphrases within this dataset has not been thoroughly analyzed. 
Guan et al.~\cite{guan_probing_2024} evaluate the performance of vision-language models under adversarial attacks without paraphrasing, targeting features important for both the language and vision encoder, reporting attack success rates up to 50\% for single-modality attacks and up to 80\% for multimodal attacks.

\textbf{Robustness of Language Models --}
Paraphrasing can critically undermine robustness in fairness-sensitive evaluations. 
Roettger et al.~\cite{roettger_political_2024} demonstrate that semantically equivalent rewordings in political compass tests substantially alter measured political bias in LLMs.
While Ceron et al.~\cite{ceron_prompt_2024} report similar effects in voting advice applications, where paraphrased or negated items lead to inconsistent outcomes. 
Even minor surface-level changes, such as punctuation, negation, or back-translated paraphrases, can severely affect model predictions~\cite{sclar_quantifying_2024,shu_you_2024,gonen_demystifying_2023}. 
Adversarial evaluation methods, such as TextFooler and BERT-Attack~\cite{jin_is_2020,li_bert_2020}, exploit synonym substitutions to mislead BERT-based classifiers with high success rates. 
Adversarial GLUE~\cite{wang_adversarial_2021} similarly shows large performance drops under systematic perturbations. 
Fairness-oriented benchmarks, such as FLEX~\cite{jung_flex_2025}, further demonstrate that adversarial manipulations can induce biased outputs.

\textbf{Robustness of Vision Models --}
Robustness in vision and vision-language models has been extensively studied, particularly in the context of retrieval and classification under distribution shifts or adversarial manipulation. 
CLIP and related models are known to be sensitive to subtle linguistic changes in textual queries, which can amplify existing biases in socially sensitive contexts~\cite{howard_socialcounterfactuals_2024}. 
Evaluation approaches include adversarial feature attacks~\cite{guan_probing_2024}, contrastive robustness testing, and controlled datasets with counterfactual demographic variations~\cite{howard_socialcounterfactuals_2024}. 

\textbf{Summary and Motivation for PRSM --}
While prior work highlights both linguistic and visual vulnerabilities, there is a lack of systematic measures that quantify semantic consistency under paraphrasing for encoder-only vision-language models. 
The Paraphrase Ranking Stability Metric (PRSM) addresses this gap by measuring the stability of global and local retrieval rankings across paraphrased queries, allowing researchers to assess fairness, reliability, and equitable deployment of multimodal systems.

\section{Measuring CLIP’s Robustness for Paraphrases}\label{methods}

We evaluate paraphrase robustness using the Social Counterfactuals dataset to control for gender, generating paraphrases of queries, and quantify stability in retrieval using our proposed measure PRSM. 

\textbf{Social Counterfactuals Dataset --}
We use the Social Counterfactuals dataset~\cite{howard_socialcounterfactuals_2024}, which provides controlled prompt sets where demographic attributes (e.g., man vs. woman) are varied in counterfactual pairs. Note that while gender is not a binary concept, the Social Counterfactual dataset only provides examples for the categories \textit{male} and \textit{female}.
This dataset allows us to evaluate not only semantic consistency under paraphrasing but also the interaction between linguistic variation and sensitive social attributes.

We generate three kinds of paraphrases: One random LLM-generated set (P1), one strategic set that paraphrases the prefix (P2), and one strategic set that paraphrases the demographic attributes (P3). Given the caption \textit{A photo of a young female academic}, the prefix is \textit{a photo of}, and attributes one and two are \textit{young} and \textit{female}.

\textbf{Generating Paraphrases --}
For the LLM-generated paraphrases (P1), we use \texttt{Llama-3.1-8B-Instruct}~\cite{grattafiori_llama_2024}. 
We use the original caption as the user input and include the following system prompt: \textit{Give me a paraphrase of the following statement, don't add information, and don't leave anything out.}. 
We sample two responses as paraphrases and manually validate a subset of 100 captions to ensure that their paraphrases are valid. 
We compare the original caption against the first paraphrase (o-c1), the original against the second paraphrase (o-c2), and all three versions (o-c1-c2).

For the prefix paraphrases, we find all unique prefixes that introduce the caption. 
There are only four: \textit{an image of, a photo of, a picture of}, and no prefix. 
We generate four versions for each caption, one for each unique prefix. 
We compare the \textit{an image of} prefix against the \textit{a picture of} prefix (p1-p2), the \textit{an image of} prefix against no prefix (p1-np), and all four versions (p1-p2-p3-np).

For the attribute paraphrase, we extract all unique values of both attributes from each caption. 
We use Open AI's GPT-5 to generate one synonym for each attribute. 
Then, we generate three paraphrases for each caption: One where we replace the first demographic attribute, one where we replace the second demographic attribute, and one where we replace both demographic attributes. 
We compare the original against paraphrasing the first attribute (o-a1), the original against paraphrasing the second attribute (o-a2), the original against paraphrasing both attributes (o-a12), and all four paraphrases (o-a1-a2-a12)

\textbf{Measuring the paraphrase robustness --}
We introduce the Paraphrase Ranking Stability Metric (PRSM) to measure the robustness of a vision-language model to paraphrased queries. 
The measure evaluates whether semantically equivalent rephrasings of a query produce consistent retrieval results.

For a given text query, PRSM first computes similarity scores between the query embedding and a set of image embeddings, producing a ranked list of images. 
The same procedure is applied independently to two paraphrased versions of the query. 
To quantify robustness, PRSM then compares the rankings of these three queries across all pairwise combinations: (original, paraphrase1), (original, paraphrase2), and (paraphrase1, paraphrase2).

The comparison can be performed in two ways:
\begin{itemize}
    \setlength\topsep{0pt}
    \setlength\partopsep{0pt}
    \setlength\itemsep{0pt}
    \setlength\itemsep{0em}
    \setlength\parskip{0em}
    \setlength\parsep{0em}
    \item \textbf{Spearman correlation} assesses \emph{global ranking stability} by measuring the correlation of the full ranked lists of images.
    \item \textbf{Top-k overlap} assesses \emph{local stability} by computing the fraction of images shared among the top-k retrieved results.
\end{itemize}

To formalize PRSM, let $Q = \{q_1, \dots, q_m\}$ be a set of paraphrased queries, and let $R(q)$ denote the ranked list of retrieved images for query $q$. We define:

\begin{itemize}
    \setlength\topsep{0pt}
    \setlength\partopsep{0pt}
    \setlength\itemsep{0pt}
    \setlength\itemsep{0em}
    \setlength\parskip{0em}
    \setlength\parsep{0em}
    \item \textbf{Global stability (Spearman correlation):}
    \[
    \text{PRSM}_{\text{global}} = \frac{2}{m(m-1)} \sum_{i<j} \rho(R(q_i), R(q_j))
    \]
    \item \textbf{Local stability (Top-k overlap):}
    \[
    \text{PRSM}_{\text{local}}(k) = \frac{2}{m(m-1)} \sum_{i<j} \frac{|R_k(q_i) \cap R_k(q_j)|}{k}
    \]
\end{itemize}

Here, $\rho$ is Spearman’s rank correlation, and $R_k(q)$ is the top-$k$ retrieved images for query $q$. 
PRSM aggregates stability across all paraphrase pairs, capturing both global ranking preservation and local retrieval consistency.

PRSM captures both global and local stability of retrieval under paraphrasing. 
High $\text{PRSM}_{\text{global}}$ indicates that the overall ordering of images is largely preserved across paraphrases.
While high $\text{PRSM}_{\text{local}}$ overlap shows that the most relevant results remain consistent.
Low PRSM values indicate that paraphrased queries yield divergent rankings, underscoring the model's sensitivity to minor linguistic variations. 
By combining these measures, PRSM provides a comprehensive view of paraphrase robustness in vision-language retrieval.

\section{Results and Discussion}

Table~\ref{tab:prsm} summarizes PRSM scores on the Social Counterfactuals dataset across the three paraphrasing strategies: (P1) LLM-generated queries, (P2) prefix-based queries, and (P3) attribute-based queries for all comparisons listed in Section~\ref{methods}.
Across all conditions, global stability is extremely low: Spearman correlations remain below 0.04, showing that semantically equivalent queries yield substantially different global rankings. 
Local stability is higher. 
For P1, top-100 overlap averages $\sim$0.64 and top-1,000 $\sim$0.73. These values are the lowest of all paraphrasing tasks and reflect the largest wording variation compared to the original captions, not only strategically targeting some sentence components, but also the entire sentence. 
P2 achieves substantially higher consistency (top-100 $\sim$0.89, top-1,000 $\sim$0.93), indicating robustness to superficial prefix changes. This robustness also holds when omitting the prefix (p1-np), indicating that the wording and even the presence of phrases that are not describing specific attribtues of an image are not relevant to the embedding and thus to downstream retrieval tasks.
P3 yields more variable results (top-100 between 0.63–0.74), suggesting sensitivity when rephrasing or combining attributes. While this is evidence for the brittleness of CLIP embeddings against paraphrasing of relevant attributes, it can also partially be an artefact of the nature of paraphrases in practice, where not every synonym conveys the exact same meaning.

\begin{table}[bth]
\centering
\begin{tabular}{m{0.5cm}l|m{1cm}m{1cm}m{1cm}|m{1cm}m{1cm}m{1cm}|m{1cm}m{1cm}m{1cm}}
\toprule
\multicolumn{1}{l}{} &  & \multicolumn{3}{c|}{\begin{tabular}[c]{@{}c@{}}Overall \\ (n=170,832)\end{tabular}} & \multicolumn{3}{c|}{\begin{tabular}[c]{@{}c@{}}Female \\ (n=72,876)\end{tabular}} & \multicolumn{3}{c}{\begin{tabular}[c]{@{}c@{}}Male \\ (n=72,876)\end{tabular}} \\
\multicolumn{1}{l}{} &  & \multicolumn{1}{c}{\begin{tabular}[c]{@{}c@{}}Spear-\\ man\end{tabular}} & \multicolumn{1}{c}{\begin{tabular}[c]{@{}c@{}}Top\\ 100\end{tabular}} & \multicolumn{1}{c|}{\begin{tabular}[c]{@{}c@{}}Top\\ 1000\end{tabular}} & \multicolumn{1}{c}{\begin{tabular}[c]{@{}c@{}}Spear-\\ man\end{tabular}} & \multicolumn{1}{c}{\begin{tabular}[c]{@{}c@{}}Top\\ 100\end{tabular}} & \multicolumn{1}{c|}{\begin{tabular}[c]{@{}c@{}}Top\\ 1000\end{tabular}} & \multicolumn{1}{c}{\begin{tabular}[c]{@{}c@{}}Spear-\\ man\end{tabular}} & \multicolumn{1}{c}{\begin{tabular}[c]{@{}c@{}}Top\\ 100\end{tabular}} & \multicolumn{1}{c}{\begin{tabular}[c]{@{}c@{}}Top\\ 1000\end{tabular}} \\
\midrule
\multirow{3}{*}{P1} & o-c1 & 0.030 & 0.642 & 0.733 & 0.040 & 0.669 & 0.763 & 0.040 & 0.688 & 0.774 \\
 & o-c2 & 0.030 & 0.641 & 0.733 & 0.040 & 0.668 & 0.762 & 0.040 & 0.688 & 0.774 \\
 & o-c1-c2 & 0.030 & 0.642 & 0.733 & 0.040 & 0.669 & 0.763 & 0.040 & 0.688 & 0.774 \\
 \midrule
\multirow{3}{*}{P2} & p1-p2 & 0.024 & 0.891 & 0.925 & 0.034 & 0.897 & 0.932 & 0.035 & 0.900 & 0.932 \\
 & p1-np & 0.023 & 0.785 & 0.852 & 0.034 & 0.800 & 0.868 & 0.035 & 0.797 & 0.861 \\
 & p1-p2-p3-np & 0.024 & 0.837 & 0.888 & 0.033 & 0.847 & 0.899 & 0.029 & 0.847 & 0.898 \\
 \midrule
\multirow{4}{*}{P3} & o-a1 & 0.021 & 0.699 & 0.772 & 0.031 & 0.723 & 0.800 & 0.032 & 0.720 & 0.788 \\
 & o-a2 & 0.025 & 0.739 & 0.811 & 0.033 & 0.787 & 0.860 & 0.037 & 0.753 & 0.819 \\
 & o-a12 & 0.023 & 0.631 & 0.726 & 0.031 & 0.684 & 0.778 & 0.029 & 0.649 & 0.739 \\
 & o-a1-a2-a12 & 0.023 & 0.682 & 0.763 & 0.031 & 0.723 & 0.807 & 0.036 & 0.701 & 0.780\\
\bottomrule
\end{tabular}
\vspace{0em}
\caption{PRSM results on the Social Counterfactuals dataset for all three paraphrasing sets. P1: LLM-generated paraphrases (o=original, c1/2=caption paraphrase 1/2), P2: strategic prefix paraphrases (p1/2/3: prefix variant 1/2/3, np: no prefix), P3: strategic attribute paraphrases (o: original, a1/2/12: paraphrasing attribute 1/2/both). Spearman reports global ranking stability and Top-$k$ reports local retrieval stability.}
\label{tab:prsm}
\vspace{-2.5em}
\end{table}

Our gender-specific analysis reveals the same overall pattern: low global stability with moderate-to-high local stability. 
Differences between male and female queries are small but consistent. Female-associated queries achieve slightly higher local overlap in P2 and P3, while male-associated queries obtain marginally higher Spearman correlations in some cases. 
These effects, though minor, indicate that paraphrase sensitivity can interact with demographic attributes, potentially leading to systematic but subtle retrieval biases that accumulate to failures across large-scale use in video retrieval applications, e.g., VBS~\cite{rossetto_interactive_2020}.

In summary, CLIP retrieval is highly unstable in global rankings and mostly at the top-$k$ level, where we find higher differences for paraphrases of high-relevance attributes that are less strategic.
This duality implies that while end users may see consistent top results, system-level representations remain fragile. 
Such instability poses risks in socially sensitive applications, where demographic fairness is critical. 
PRSM provides an explicit diagnosis of this behavior, revealing robustness to shallow variation but fragility under semantic reformulation.

\section{Conclusion}

We introduced the PRSM as a measure to evaluate the robustness of vision-language models under paraphrastic variation. 
Using the Social Counterfactuals dataset, we conducted a case study on gender-specific queries and found that CLIP exhibits low global stability, as measured by Spearman correlation, but moderate local stability in top-$k$ retrieval. We also find small but consistent gender differences.
These results reveal an important tension: while user-facing retrieval outcomes may appear stable, the underlying global rankings are highly sensitive to minor linguistic changes. 
Such instability poses risks for fairness and reliability, particularly in socially sensitive applications, for instance, in law enforcement or healthcare.

PRSM contributes a systematic way to quantify these vulnerabilities by combining global and local perspectives on ranking stability. 
Our findings underscore the need for paraphrase-invariant training methods or post-hoc methods for mitigation and evaluation practices that explicitly account for linguistic variation. 

Our analysis provides initial results on the robustness of CLIP embeddings for paraphrases; however, we only report results for one dataset and separate the findings by only one demographic attribute. 
Further research would expand to include more datasets and more demographic or other social attributes, such as race, religion, or disability status. 
Based on the findings of previous work on the brittleness of LLMS~\cite{ceron_prompt_2024,roettger_political_2024}, another important direction for future work is the extension to autoregressive vision-language models (BLIP~\cite{li_blip_2022}) and the mitigation of possible biases in them, similar to Lan et al.~\cite{lan_my_2025}.

\bibliographystyle{splncs04}
\bibliography{mmm2025paper}
%




\end{document}